\documentclass[letterpaper, 10 pt, conference]{ieeeconf}  

\IEEEoverridecommandlockouts                              

\usepackage{amsmath,amsfonts,bm}

\def\eqref#1{equation~\ref{#1}}

\def\1{\bm{1}}

\DeclareMathAlphabet{\mathsfit}{\encodingdefault}{\sfdefault}{m}{sl}
\SetMathAlphabet{\mathsfit}{bold}{\encodingdefault}{\sfdefault}{bx}{n}

\DeclareMathOperator*{\argmax}{arg\,max}
\DeclareMathOperator*{\argmin}{arg\,min}

\usepackage{graphics} 
\usepackage{epsfig} 
\usepackage{mathptmx} 
\usepackage{times} 
\usepackage{amsmath} 
\usepackage{amssymb}  

\usepackage{mathtools}

\usepackage{multirow, booktabs}

\usepackage{xcolor}

\usepackage[utf8]{inputenc} 
\usepackage{pifont}
\newcommand{\cmark}{\ding{51}}%
\newcommand{\xmark}{\ding{55}}%

\usepackage{subfigure}
\usepackage{caption}
\usepackage{tabularx}
\usepackage{float}
\usepackage{epsfig}
\usepackage{epstopdf}

\usepackage{algorithm}
\usepackage[noend]{algpseudocode}
\usepackage{multirow}
\usepackage[section]{placeins}
\usepackage{enumerate}

\newcommand{\pname}{BaSiL}

\title{\LARGE \bf
Streaming LifeLong Learning With Any-Time Inference
}

\author{Soumya Banerjee$^{1}$, Vinay Kumar Verma$^{2}$, and Vinay P. Namboodiri$^{3}$
\thanks{$^{1}$ IIT Kanpur, India,
        {\tt\small soumyab@cse.iitk.ac.in}}
\thanks{$^{2}$ Duke University, USA,
        {\tt\small vinaykumar.verma@duke.edu}}
\thanks{$^{3}$ University Of Bath, UK,
        {\tt\small vpn22@bath.ac.uk}}
}

\begin{document}

\maketitle
\thispagestyle{empty}
\pagestyle{empty}

\begin{abstract}

Despite rapid advancements in lifelong learning (LLL) research, a large body of research mainly focuses on improving the performance in the existing \textit{static} continual learning (CL) setups. These methods lack the ability to succeed in a rapidly changing \textit{dynamic} environment, where an AI agent needs to quickly learn new instances in a `single pass' from the non-i.i.d (also possibly temporally contiguous/coherent) data streams without suffering from catastrophic forgetting. For practical applicability, we propose a novel lifelong learning approach, which is streaming, i.e., a single input sample arrives in each time step, single pass, class-incremental, and subject to be evaluated at any moment. To address this challenging setup and various evaluation protocols, we propose a Bayesian framework, that enables fast parameter update, given a single training example, and enables any-time inference. We additionally propose an implicit regularizer in the form of snap-shot self-distillation, which effectively minimizes the forgetting further. We further propose an effective method that efficiently selects a subset of samples for online memory rehearsal and employs a new replay buffer management scheme that significantly boosts the overall performance. Our empirical evaluations and ablations demonstrate that the proposed method outperforms the prior works by large margins.

\end{abstract}


\section{INTRODUCTION}

\begin{figure}[t]
        \centering
        \includegraphics[width=8cm, height=2.6cm]{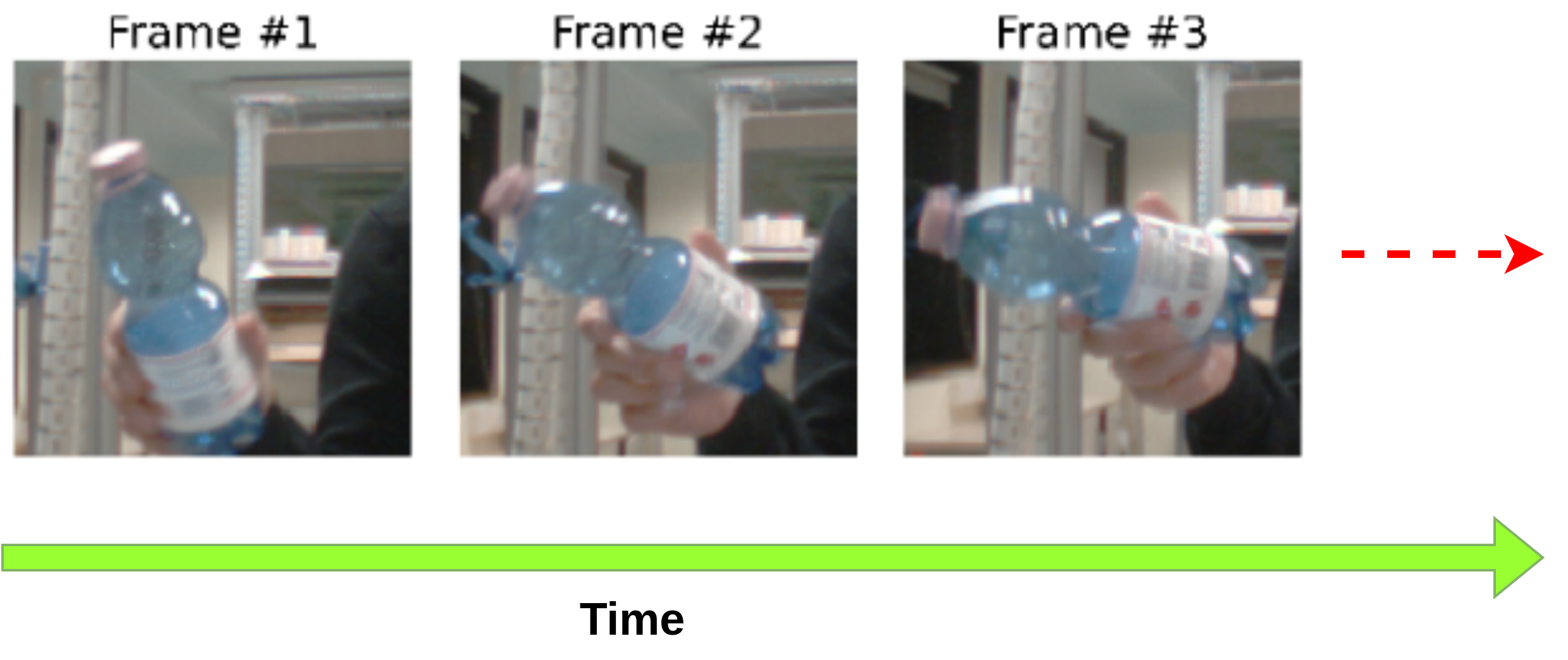}

        \vspace{-0.6em}

        \caption{SL addresses the problem of learning continuously from a stream of temporally correlated (non-i.i.d) labeled data, for e.g., learning to recognize an object from multiple viewpoints with no forgetting. In this fig., 3 temporally correlated frames of bottle are taken from iCubWorld 1.0~\cite{fanello2013icub}.} 
        \label{fig:icubworld_streaming_learning_example}
        \vspace{-1.7em}
\end{figure}

In this paper, we aim to achieve `any-time-inference' ability in \emph{streaming lifelong learning} (SL). This is an appealing ability as it allows practical applicability of AI agents/robots, which can be continually updated and deployed without catastrophic forgetting~\cite{french1999catastrophic}. While deep learning models can achieve state-of-the-art performance throughout various tasks ranging across vision, language, speech, etc., these models suffer from catastrophic forgetting~\cite{mccloskey1989catastrophic}, if trained incrementally with sequentially arriving data. LifeLong Learning (LLL) or continual learning (CL)~\cite{kirkpatrick2017overcoming} studies the problem of training a deep neural network (DNN) continuously with sequentially coming data without forgetting. This paper introduces a novel approach, which enables DNN for class-incremental streaming classification. \emph{Class-incremental streaming classification/learning} (CISL) is a challenging variant of LLL~\cite{hayes2019memory,hayes2019remind}, with the following essential and desirable properties:
\begin{enumerate}[i)]
        
        \item Learner sees each labeled instance only once (\textit{desirable}),
        
        \item Learner needs to adapt the newly available data immediately (\textit{desirable}),
        
        \item The input data stream can be non-i.i.d and temporally contiguous/coherent (\textit{essential}),
        
        \item The AI agent can be evaluated at any moment, i.e., `any-time-inference' (\textit{essential}),
        
        \item The AI agent should be able to predict a class label among all the observed classes so far during inference without having the task-id~\cite{chaudhry2018riemannian} (\textit{essential}),
        
        \item For practical applicability, learner must limit their memory usage (\textit{desirable}).

\end{enumerate}
A wide variety of CL approaches~\cite{kirkpatrick2017overcoming,aljundi2018memory,nguyen2017variational,rios2018closed,rebuffi2017icarl,zenke2017continual,mallya2018packnet} have been proposed to alleviate catastrophic forgetting in DNN. However, most of the existing approaches mainly focus on `incremental-batch-learning' (IBL)~\cite{kirkpatrick2017overcoming}, where the input data in each incremental step arrives in batches, and the agent visits them multiple times in order to enable CL. While these methods are applicable in a static environment, these methods are ill-suited for a rapidly changing dynamic environment, where an AI agent requires to learn from the newly available knowledge immediately with no forgetting. Different from this, `streaming learning' (SL) imposes a unique set of restrictions/challenges (as mentioned earlier) and has not received widespread attention like the other existing CL approaches. However, its utility is apparent, as it enables the practical deployment of autonomous AI agents. Hayes et al.~\cite{hayes2019remind} argued that SL is closer to biological learning than other existing CL scenarios. Fig.~\ref{fig:icubworld_streaming_learning_example} demonstrates an example of streaming classification scenario. In Table~\ref{Table_baseline_categorization}, we classify the existing CL approaches in accordance to the underlying assumption that they impose. It can be observed that ExStream~\cite{hayes2019memory} and REMIND~\cite{hayes2019remind} are the only two CL approaches, that focus on streaming classification. However, it is worth mentioning that ExStream~\cite{hayes2019memory} uses full buffer replay, which violates the subset buffer replay constraint of CISL. Furthermore, while REMIND~\cite{hayes2019remind} satisfies all the CISL constraints, it stores a comparably large number of past samples in memory, which limits its applicability. 


\begin{table*}[ht] 
\scriptsize
\centering
\caption{Categorization of the baseline approaches  depending on the underlying simplifying assumptions they impose. In $\zeta(n)$, $n$ represents the number of gradient steps required to train the corresponding model. $\zeta(n) \gg \zeta(1)$. `-' indicates, that we are unable to find the exact value.}
\label{Table_baseline_categorization}

\vspace{-1em}

\resizebox{\textwidth}{!}{

\small


\begin{tabular}{ c  c  c c  c c c  c c c c  c c c } 
        \toprule

        &  &  &  & \multicolumn{6}{c}{\textbf{{`Class-Incremental Streaming Learning'} (CISL) Crucial Properties}} &  &  &   &  \\

        \cmidrule{5-10}
        
        \shortstack{{Methods}} & \shortstack{{Type}} & \shortstack{{Bayesian} \\ {Framework}} & \shortstack{{Batch-Size} \\ $(N_{t})$} & \shortstack{{Fine-tunes}} & \shortstack{{Single Pass} \\ {Learning}} & \shortstack{{CIL}}  &  \shortstack{{Subset} \\ {Buffer Replay}} & \shortstack{{Training} \\ {Time}} & \shortstack{Inference \\ Time} & \shortstack{Violates Any \\ {\emph{CISL} Constraint}} & \shortstack{{Memory} \\ {Capacity}} & \shortstack{{Regularization} \\ {Based}}  & \shortstack{{Memory} \\ {Based}} \\
        
        \midrule
        
        {EWC~\cite{kirkpatrick2017overcoming}} & Batch & \xmark & $N_{t} \gg 1$ & \xmark & \xmark & \xmark & n/a & $\zeta(n)$ & $\zeta(1)$ & \textcolor{red}{\cmark} & n/a & \cmark & \xmark \\

        \midrule
        
        {MAS~\cite{aljundi2018memory}} & Batch & \xmark & $N_{t} \gg 1$ & \xmark & \xmark & \xmark & n/a & $\zeta(n)$ & $\zeta(1)$ & \textcolor{red}{\cmark} & n/a & \cmark & \xmark \\

        \midrule


        
        {VCL~\cite{nguyen2017variational}} & Batch & \cmark & $N_{t} \gg 1$ & \xmark & \xmark & \xmark & n/a & $\zeta(n)$ & $\zeta(1)$ & \textcolor{red}{\cmark} & n/a & \cmark & \xmark \\

        \midrule
        
        {Coreset VCL~\cite{nguyen2017variational}} & Batch & {\cmark} & $N_{t} \gg 1$ & \cmark & \xmark & \xmark & \xmark & $\zeta(n)$ & $\zeta(n)$ & \textcolor{red}{\cmark} & - & \cmark & \cmark \\

        

        \midrule
        
        {GDumb~\cite{prabhu2020gdumb}} & Online & {\xmark} & $N_{t} \gg 1$ & \cmark & \xmark & \cmark & \xmark & $\zeta(1)$ & $\zeta(n)$ & \textcolor{red}{\cmark} & - & \xmark & \cmark \\

        \midrule
        
        {TinyER~\cite{chaudhry2019tiny}} & Online & \xmark & $N_{t} \gg 1$ & \xmark & \cmark & \cmark & \cmark & $\zeta(1)$ & $\zeta(1)$ & \xmark & $\leq 5 \%$ & \xmark & \cmark \\

        \midrule

        {DER~\cite{buzzega2020dark}} & Online & \xmark & $N_{t} \gg 1$ & \xmark & \cmark & \cmark & \cmark & $\zeta(1)$ & $\zeta(1)$ & \xmark & $\leq 5 \%$ & \cmark & \xmark \\

        \midrule

        {DER++~\cite{buzzega2020dark}} & Online & \xmark & $N_{t} \gg 1$ & \xmark & \cmark & \cmark & \cmark & $\zeta(1)$ & $\zeta(1)$ & \xmark & $\leq 5 \%$ & \cmark & \cmark \\

        \midrule
        






        
        {ExStream~\cite{hayes2019memory}} & Streaming & \xmark & $N_{t} = 1$ & \xmark & \cmark & \cmark & \xmark & $\zeta(1)$ & $\zeta(1)$ & \textcolor{red}{\cmark} & $\leq 5 \%$ & \xmark & \cmark \\

        \midrule
        
        
        {REMIND~\cite{hayes2019remind}} & Streaming & \xmark & $N_{t} = 1$ & \xmark & \cmark & \cmark & \cmark & $\zeta(1)$ & $\zeta(1)$ & \xmark & $\gg 10 \%$ & \xmark & \cmark \\
        
        \midrule
        
        {\textbf{Ours}} & Streaming & \cmark & $N_{t} = 1$ & \xmark & \cmark & \cmark & \cmark & $\zeta(1)$ & $\zeta(1)$ & \xmark & $\leq 5 \%$ & \cmark & \cmark \\

        \midrule

        \multicolumn{1}{c}{\textbf{CISL Constraints}} & & & $N_{t} = 1$   & \xmark & \cmark & \cmark & \cmark & $\zeta(1)$ & $\zeta(1)$ &  &  &   &  \\

        \bottomrule
\end{tabular}

}

\vspace{-2.2em}

\end{table*}


In this work, we overcome the aforementioned limitations in a principled manner and introduce a novel CISL method, dubbed as \textbf{Ba}yesian Class Incremental \textbf{S}tream\textbf{i}ng \textbf{L}earning ({\pname}), that seeks to incorporate the ability of rehearsal with a functional regularization into a simple yet effective framework of streaming variational Bayes. We leverage both the replay samples and the sequential data to compute the joint likelihood in order to approximate the posterior in each incremental step. As a consequence, the proposed approach does not require explicit finetuning unlike approaches like~\cite{prabhu2020gdumb}, and can be evaluated `on-the-fly' at any moment. We also propose a novel online `loss-aware' buffer management policy and various sample selection strategies that include the most informative sample(s) in memory or select for rehearsal, which significantly boosts the model's performance. Our experimental results on three benchmark datasets demonstrate the effectiveness of our proposed method to circumvent forgetting in highly challenging streaming scenarios. 

\textbf{Our contributions can be summarized as follows:}
\begin{itemize}

        \item We introduce {\pname}, a novel rehearsal-based dual regularization framework, which includes a streaming Bayesian framework and a functional regularizer, to alleviate forgetting in challenging CISL setup,
        
        \item We propose a novel online `loss-aware' replay buffer management policy and various sampling strategies which significantly boosts the model's performance,
        
        \item  We empirically show that the proposed method can be evaluated at any time without requiring to finetune unlike ~\cite{prabhu2020gdumb,nguyen2017variational} with no significant drop in final accuracy,
        
        \item Empirical evaluations and ablations on three benchmark datasets validate the superiority of {\pname} over the existing baselines.

\end{itemize}


\section{RELATED WORK}

Parameter-isolation-based approaches train different subsets of model parameters on sequential tasks. PNN~\cite{rusu2016progressive}, DEN~\cite{yoon2017lifelong} expand the network to accommodate the new task. PathNet~\cite{fernando2017pathnet}, PackNet~\cite{mallya2018packnet}, Piggyback~\cite{mallya2018piggyback}, and HAT~\cite{serra2018overcoming} train different subsets of network parameters on each task.

Regularization-based approaches use an extra regularization term in the loss function to enable continual learning. LWF~\cite{li2017learning} uses knowledge distillation~\cite{hinton2015distilling} loss to prevent forgetting. EWC~\cite{kirkpatrick2017overcoming}, IMM~\cite{lee2017overcoming}, SI~\cite{zenke2017continual} and MAS~\cite{aljundi2018memory} regularize by penalizing changes to the important weights of the network.

Rehearsal-based approaches replay a subset of past training data during sequential learning. iCaRL~\cite{rebuffi2017icarl}, SER~\cite{isele2018selective}, and TinyER~\cite{chaudhry2019tiny} use memory replay when training on a new task. DER/DER++~\cite{buzzega2020dark} uses knowledge distillation~\cite{hinton2015distilling} and memory replay while learning a new task. DGR~\cite{shin2017continual}, MeRGAN~\cite{wu2018memory}, and CloGAN~\cite{rios2018closed} uses generative models to replay past samples during continual learning.

VCL~\cite{nguyen2017variational} leverages Bayesian inference to mitigate forgetting~\cite{mccloskey1989catastrophic}. However, the approach, when na\"ively adapted, performs poorly in SL. Additionally, it requires task-id during inference. Furthermore, explicit finetuning with the coreset samples restricts it from performing any time inference.

REMIND~\cite{hayes2019remind} is a recently proposed rehearsal-based SL approach, which follows a setting close to the one proposed. However, it stores a large number of past examples compared to the other baselines; for e.g., iCaRL~\cite{rebuffi2017icarl} stores 10K past examples for ImageNet experiment, whereas REMIND stores 1M past examples. Furthermore, it actually uses a lossy compression to store past samples, which is merely an engineering technique, not an algorithmic improvement, and can be used by any CL approach.

\section{PROPOSED APPROACH}

\subsection{Streaming Learning With A Single Example}\label{streaming_learning_with_a_single_example}

\begin{figure}[t]
  \centering
  \includegraphics[width=6.5cm, height=2.8cm]{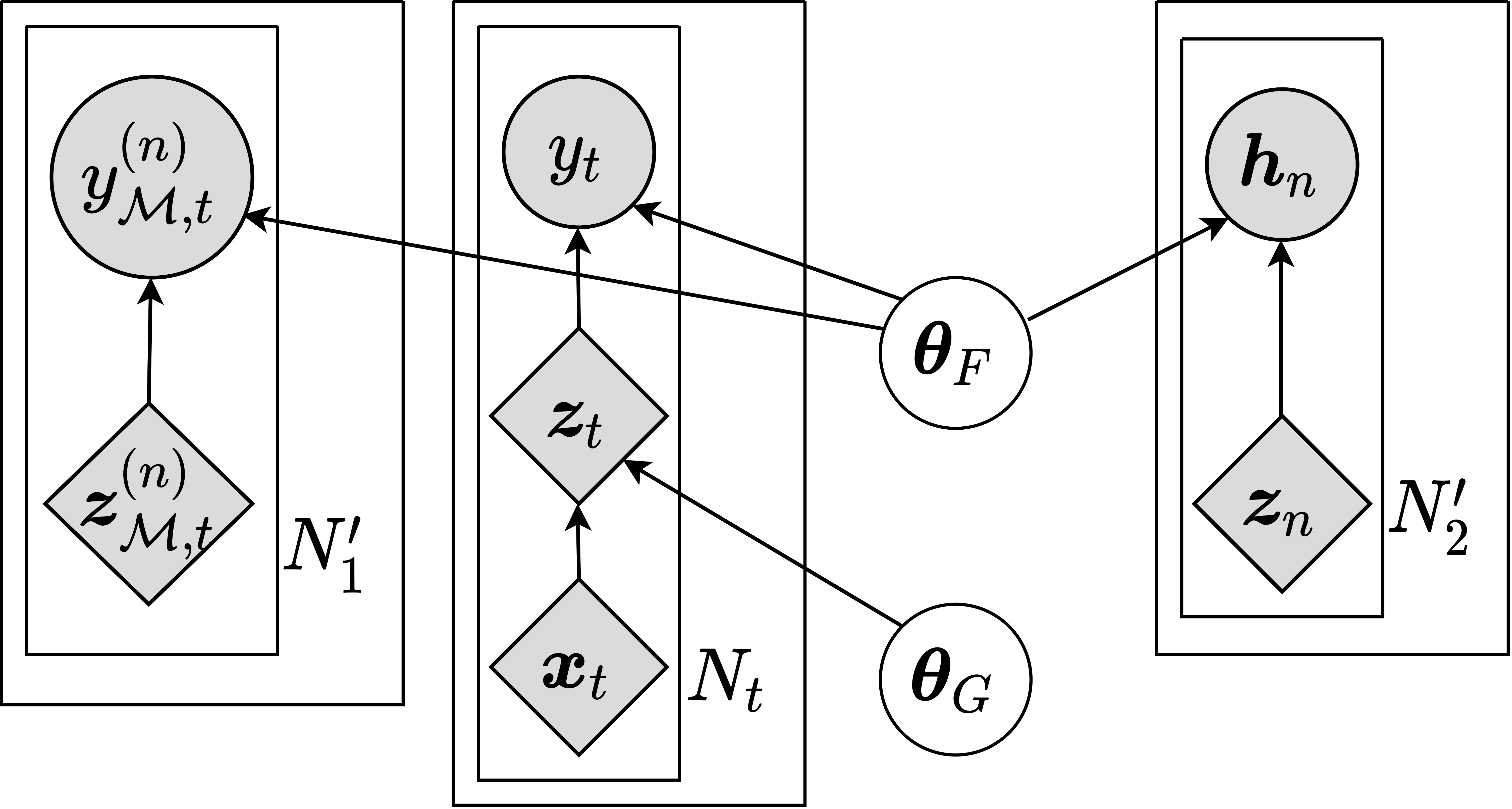}
  
  \vspace{-0.5em}

  \caption{Schematic representation of the proposed model. $\boldsymbol{\theta}_{G}$ denotes the parameters of the frozen feature extractor $G(\cdot)$, whereas $\boldsymbol{\theta}_{F}$ denotes the parameters of BNN $F(\cdot)$.} 
  \label{fig:proposed_streaming_learning_model}
  \vspace{-1.8em}
\end{figure}

In the following, we introduce {\pname}, which trains a convolutional neural network (CNN) in CISL setup. Formally, we separate the CNN into two networks (Fig.~\ref{fig:proposed_streaming_learning_model}): $(i)$ non-plastic feature extractor $G(\cdot)$, containing the initial $U$ layers of CNN, and $(ii)$ plastic neural network $F(\cdot)$, containing the final $V$ layers. For a given input image $\boldsymbol{x}$, the predicted class label is computed as: $y = F(G(\boldsymbol{x}))$. We initialize the parameters of feature extractor $G(\cdot)$ with the pre-trained weights and keep it frozen throughout SL. We model the plastic network $F(\cdot)$ as a \emph{Bayesian-neural-network} (BNN)~\cite{neal2012bayesian,jospin2020hands} and optimize its parameters with the sequentially coming data in CISL setup without forgetting.

In the below, we describe how the plastic network $F(\cdot)$ (BNN) is optimized with a single posterior approximation in each incremental step, $t$, with a single data point: $\boldsymbol{D}_{t} = \left\{ {d}_{t} \right\} = \left\{ \left(\boldsymbol{x}_{t}, y_{t}\right) \right\}$, arriving sequentially in a streaming manner with no forgetting~\cite{mccloskey1989catastrophic}.

\textbf{Variational Posterior Estimation.} Streaming learning naturally emerges from the Bayes' rule~\cite{broderick2013streaming}; given the posterior $p(\boldsymbol{\theta} | \boldsymbol{D}_{1:{t - 1}})$, whenever a new data: $\boldsymbol{D}_{t} = \{ {d}_{t} \} = \{(\boldsymbol{x}_{t}, y_{t})\}$ arrives, the new posterior $p(\boldsymbol{\theta} | \boldsymbol{D}_{1:t})$ can be computed by combining the previous posterior and the new data likelihood, i.e., $p(\boldsymbol{\theta} | \boldsymbol{D}_{1:t}) \propto p(\boldsymbol{D}_{t} | \boldsymbol{\theta}) \; p(\boldsymbol{\theta} | \boldsymbol{D}_{1:{t - 1}})$, where the old posterior is treated as the prior. However, for any complex model, the exact Bayesian inference is not tractable, and an approximation is needed. A Bayesian neural network (BNN)~\cite{neal2012bayesian,blundell2015weight} commonly approximates the posterior with a variational posterior $q(\boldsymbol{\theta})$ by minimizing the KL divergence (in Eq.~\ref{eq:1}), or equivalently by maximizing the evidence lower bound (ELBO) (in Eq.~\ref{eq:2}):
\begin{align}\label{eq:1}
    \boldsymbol{L}_{t}(\boldsymbol{\theta}) &= \argmin_{q \in {Q}} \text{KL} \left( q_{t}(\boldsymbol{\theta}) \left|\right| \frac{1}{Z_{t}} q_{t-1}(\boldsymbol{\theta}) p(\boldsymbol{D}_{t}|\boldsymbol{\theta}) \right) \\
  \begin{split}\label{eq:2}
    &\backsimeq \argmax_{q \in {Q}} \mathbb{E}_{\boldsymbol{\theta} \sim q_{t}(\boldsymbol{\theta})}\left[ \log p(\boldsymbol{D}_{t}|\boldsymbol{\theta}) \right] - \text{KL} \left( q_{t}(\boldsymbol{\theta}) \left| \right| q_{t-1}(\boldsymbol{\theta}) \right)
  \end{split}
\end{align}
KL-divergence in Eq.~\ref{eq:2} works as an implicit parameter regularizer, by keeping the prior and the posterior distribution close to each other, which can minimize forgetting in sequential learning. However, there are a few concerns. Firstly, optimizing the ELBO (in Eq.~\ref{eq:2}) to approximate the posterior $p(\boldsymbol{\theta}|\boldsymbol{D}_{1:t})$ with only a single training example, i.e., $\boldsymbol{D}_{t} = \left\{ {d}_{t} \right\} = \left\{ \left(\boldsymbol{x}_{t}, y_{t}\right) \right\}$, in each incremental step, $t$, can fail in SL~\cite{ghosh2018structured}. Furthermore, likelihood estimation from a single example: $\boldsymbol{D}_{t} = \left\{ {d}_{t} \right\} = \left\{ \left(\boldsymbol{x}_{t}, y_{t}\right) \right\}$ can bias the model towards the new data disproportionately and can maximize the confusion during inference in class incremental learning (CIL) setup~\cite{chaudhry2018riemannian}. 


We, therefore, propose to estimate the likelihood from both the new example: $\boldsymbol{D}_{t}$ and the previously observed examples in order to approximate the new posterior: $p(\boldsymbol{\theta}|\boldsymbol{D}_{1:t})$. For this purpose, a small fraction $(\leq 5\%)$ of past observed examples are stored as representatives in a \emph{fixed-sized} tiny episodic memory buffer $\boldsymbol{M}$. However, instead of storing raw pixels (images) $\boldsymbol{x}$, we store the embedding $\boldsymbol{z} = G(\boldsymbol{x})$, where $\boldsymbol{z} \in \mathbb{R}^{d}$. This allows us to avoid the cost of recomputing the image-embeddings and expedites the learning even further. It also saves a significant amount of space and allows us to keep more examples in a small budget.

During training on the new example: $\boldsymbol{D}_{t} = \left\{ {d}_{t} \right\} = \left\{ \left(\boldsymbol{x}_{t}, y_{t}\right) \right\}$, we select a subset of samples: $\boldsymbol{D}_{\boldsymbol{M}, t}$ from the memory $\boldsymbol{M}$, where: $(i)\boldsymbol{D}_{\boldsymbol{M}, t} \subset \boldsymbol{M}, \quad (ii) |\boldsymbol{D}_{\boldsymbol{M}, t}| = {N}^{\prime}_{1} \ll |\boldsymbol{M}|$, \emph{instead of replaying the whole buffer}, and compute the likelihood jointly with the new example $\boldsymbol{D}_{t}$ to estimate the new posterior. We, therefore, compute the new posterior as follows: $p(\boldsymbol{\theta} | \boldsymbol{D}_{1:t}) \propto p(\boldsymbol{D}_{t} | \boldsymbol{\theta}) \; p(\boldsymbol{D}_{\boldsymbol{M}, t} | \boldsymbol{\theta}) \; p(\boldsymbol{\theta} | \boldsymbol{D}_{1: {t - 1}})$. Since the exact Bayesian inference is intractable, we approximate it with a variational posterior $q_{t}(\boldsymbol{\theta})$ as follows:   
\begin{align}\label{eq:3}
    \boldsymbol{L}^{1}_{t}(\boldsymbol{\theta}) = \argmin_{q \epsilon {Q}} \text{KL}\left( q_{t}(\boldsymbol{\theta}) \left|\right| \frac{1}{Z_{t}} q_{t - 1}(\boldsymbol{\theta}) p(\boldsymbol{D}_{t} | \boldsymbol{\theta}) p(\boldsymbol{D}_{\boldsymbol{M}, t} | \boldsymbol{\theta})  \right)
\end{align}
\emph{Note that Eq.~\ref{eq:3} consists of two likelihood terms, that is, one for the new data and other for the subset buffer replay. It minimizes the confusion in the network parameters while predicting a class label over all the observed classes in CIL setup. Further, it allows the model to be evaluated `on-the-fly' at any moment.}


The above minimization (in Eq.~\ref{eq:3}) can be equivalently written as the maximization of the evidence lower bound (ELBO) as follow:
\begin{align}\label{eq:4}
    \boldsymbol{L}^{1}_{t}(\boldsymbol{\theta}) &= \begin{multlined}[t]
    \mathbb{E}_{\boldsymbol{\theta} \sim q_{t}(\boldsymbol{\theta})} \left[ \log{p(y_{t} | \boldsymbol{\theta}, G(\boldsymbol{x}_{t}))} \right] \\
    + \sum_{n = 1}^{N_{1}^{\prime}} \mathbb{E}_{\boldsymbol{\theta} \sim q_{t}(\boldsymbol{\theta})} \left[ \log{p(y_{\boldsymbol{M}, t}^{(n)} | \boldsymbol{\theta}, \boldsymbol{z}_{\boldsymbol{M}, t}^{(n)})} \right] \\
    - \lambda_{1} \cdot \text{KL}\left(q_{t}(\boldsymbol{\theta}) \left|\right| q_{t - 1}(\boldsymbol{\theta}) \right)
    \end{multlined}
\end{align}
where: $(i)$ $\boldsymbol{D}_{t} = \{ {d}_{t} \} = \{ ( \boldsymbol{x}_{t}, y_{t} ) \}$, $(ii)$ $\boldsymbol{D}_{\boldsymbol{M}, t} = \{ {d}^{(n)}_{\boldsymbol{M}, t} \}^{N_{1}^{\prime}}_{n = 1} = \{ ( \boldsymbol{z}_{\boldsymbol{M}, t}^{(n)}, y_{\boldsymbol{M}, t}^{(n)} ) \}^{N_{1}^{\prime}}_{n = 1}$, $(iii)$ $\boldsymbol{D}_{\boldsymbol{M}, t} \subset \boldsymbol{M}$, $(iv)$ $| \boldsymbol{D}_{\boldsymbol{M}, t} | = N_{1}^{\prime} \ll | \boldsymbol{M} |$, and $(v)$ $\lambda_{1}$ is a hyper-parameter.



\textbf{Snap-Shot Self Distillation.} It is worth noting that KL divergence minimization (in Eq.~\ref{eq:4}) between prior and posterior distribution  works as an inherent parameter regularizer, and minimizes the changes in the network parameters in SL. However, its effect may weaken over time due to the presence of distribution shift and temporal coherence in the input data-stream. Furthermore, the initialization of the prior with the old posterior at each incremental step can introduce information loss in the network for a longer sequence of SL. On these grounds, we propose a functional regularizer, which encourages the network to mimic the output responses as produced in the past for the previously observed samples. Specifically, we propose to minimize KL divergence between class-probability scores obtained in past and current incremental step $t$. In particular, we uniformly select $N^{\prime}_{2}$ samples along with their logits and minimize the objective in Eq.~\ref{eq:5}, or equivalently under mild assumptions of knowledge-distillation~\cite{hinton2015distilling}, we minimize the mean-squared-error between the corresponding logits as in Eq.~\ref{eq:6}:
\vspace{-0.75em}
\begin{align}\label{eq:5}
    \boldsymbol{L}^{2}_{t}(\boldsymbol{\theta}) &= \lambda_{2} \cdot \sum^{N_{2}^{\prime}}_{j = 1} \mathbb{E}_{\boldsymbol{\theta} \sim q_{t}(\boldsymbol{\theta})} \left[ \text{KL} \left( \text{softmax}(\boldsymbol{h}_{j}) \; \left|\right| \; \text{softmax}(f_{\boldsymbol{\theta}}(\boldsymbol{z}_{j})) \right) \right] \\
  \begin{split}\label{eq:6}
    &= \lambda_{2} \cdot \sum^{N_{2}^{\prime}}_{j = 1} \mathbb{E}_{\boldsymbol{\theta} \sim q_{t}(\boldsymbol{\theta})} \left[ \left|\left| \boldsymbol{h}_{j} - f_{\boldsymbol{\theta}}(\boldsymbol{z}_{j}) \right|\right|^{2}_{2} \right]
  \end{split}
\end{align}
where: $(i)$ $f(\cdot)$ denotes plastic network without softmax activation, $(ii)$ $\lambda_{2}$: hyper-parameter, $(iii)$ $\boldsymbol{h}_{j}$ denotes logits.

One important highlight of Eq.~\ref{eq:6} is that the stored logits are updated in an online manner. That is, we replace the stored logits with the newly predicted logits whenever a sample is used for rehearsal. Therefore, it is called \emph{snap-shot self-distillation}~\cite{yang2019snapshot}. It essentially prevents the model from being constrained to match the sub-optimal initial logits and minimizes the information loss in the network further.

\vspace{-0.3em}

\subsection{Informative Past Sample Selection For Replay}\label{informative_past_sample_selection_for_replay}

\vspace{-0.2em}

We consider the following strategies for selecting past informative samples for memory replay:

\textbf{Uniform Sampling (Uni).} In this approach, samples are selected uniformly random from memory. 


\textbf{Uncertainty-Aware Positive-Negative Sampling (UAPN).} UAPN selects ${N_{1}^{\prime}}/{2}$ samples with the highest uncertainty scores (negative samples) and ${N_{1}^{\prime}}/{2}$ samples with the lowest uncertainty scores (positive samples). 



\textbf{Loss-Aware Positive-Negative Sampling (LAPN).} LAPN selects ${N_{1}^{\prime}}/{2}$ samples with the highest loss-values (negative-samples), and ${N_{1}^{\prime}}/{2}$ samples with the lowest loss-values (positive-samples). 


\vspace{-0.4em}

\subsection{Memory Buffer Replacement Policy}\label{memory_buffer_replacement_policy}

\vspace{-0.2em}

\begin{table}[t]


\scriptsize
\centering
\caption{Memory buffer capacity used for various datasets.}
\label{Table_buffer_capacity}

\vspace{-0.75em}


\scriptsize

\begin{tabular}{ c | c | c | c } 
  \toprule
    
  \textbf{Dataset} & \shortstack{ImageNet100} & \shortstack{iCubWorld 1.0} & \shortstack{CORe50}\\
    
  \midrule
  
    \shortstack{\textbf{Buffer}  \textbf{Capacity}} & 1000 & 180 & 1000 \\
    
    \midrule
   
    \shortstack{\textbf{Training-Set} \textbf{Size}} & 127778 & 6002 & 119894 \\
   
 \bottomrule
\end{tabular}


\vspace{-2.2em}

\end{table}




In this work, we consider the following strategies to replace a stored example with a newly arriving sample.



\textbf{Loss-Aware Weighted Class Balancing Replacement (LAWCBR).} In this approach, whenever the buffer is full, we remove a sample from the majority class, i.e., $y_{r} = \argmax \text{ClassCount}(\boldsymbol{M})$. We weigh each sample of the majority class inversely w.r.t their loss, i.e., $w^{y_{r}}_{i} \propto \frac{1}{l^{y_{r}}_{i}}$ and use these weights as the replacement probability; the lesser the loss, the more likely to be removed.


\textbf{Loss-Aware Weighted Random Replacement With A Reservoir (LAWRRR).}   In this approach, we propose a novel variant of reservoir sampling~\cite{vitter1985random} to replace an existing sample whenever the buffer is full. We weigh each stored sample inversely w.r.t the loss, i.e., $w_{i} \propto \frac{1}{l_{i}}$, and proportionally to the number of examples representative of that particular sample's class, i.e., $w_{i} \propto \text{ClassCount}(\boldsymbol{M}, y_{i})$. We combine these two scores and use that as the replacement probability; the higher the weight, the more likely to be replaced.

\vspace{-0.4em}

\subsection{Efficient Buffer Update}\label{efficient_buffer_update}

\vspace{-0.2em}


Loss/uncertainty-aware sampling strategies require computing these quantities in each incremental step for all the stored examples. However, it becomes computationally expensive with the larger replay buffer size. We, therefore, store the corresponding loss-values and uncertainty-scores along with the examples in the replay buffer. Since these quantities are just scalar values, the additional storage cost is negligible, but it makes the learning pipeline computationally efficient. Every time a sample is selected for memory replay, its loss and uncertainty are replaced by the new values. Additionally, we also replace the stored logits with the new logits. Empirically, we observe that the model's accuracy degrades if we don't update the logits with the new logits.


\subsection{Training}\label{training}

Training the plastic network (BNN) $F(\cdot)$ requires specification of $q(\boldsymbol{\theta})$ and, in this work, we model $\boldsymbol{\theta}$ by stacking up the parameters (weights \& biases) of the network $F(\cdot)$. We use a Gaussian mean-field posterior $q_{t}(\boldsymbol{\theta})$ for the network parameters, and choose the prior distribution, i.e., $q_{0}(\boldsymbol{\theta}) = p(\boldsymbol{\theta})$, as multivariate Gaussian distribution. We train the network $F(\cdot)$ by maximizing the ELBO in Eq.~\ref{eq:4} and minimizing the mean-squared-error in Eq.~\ref{eq:6}. For memory replay in Eq.~\ref{eq:4}, we select past informative samples using the strategies mentioned in Section~\ref{informative_past_sample_selection_for_replay}, and we use \emph{uniform sampling} to select samples from memory to be used in Eq.~\ref{eq:6}. Fig.~\ref{fig:proposed_streaming_learning_model} shows the schematic diagram of the proposed model as well as the learning process.

\vspace{-0.3em}

\section{Experiments}\label{experiments}

\vspace{-0.2em}

\subsection{Datasets And Data Orderings}\label{datasets_and_data_orderings}


\begin{table*}[!htbp]
  \scriptsize
  \centering
  
  \caption{$\boldsymbol{\Omega}_{\text{all}}$ results with their associated standard deviations. For each experiment, the method with best performance in \emph{`streaming-learning-setup'} is highlighted in \textbf{Bold}. The reported results are averaged over $10$ runs with different permutations of the data. Offline model is trained only once. ${\widehat{\text{Offline}} =  \frac{1}{T}{\sum_{t = 1}^{T}{\boldsymbol{\alpha}_{\text{offline}, t}}}}$, where $T$ is the total number of testing events. `-' indicates
  experiments we are unable to run, because of compatibility issues.}
  \label{Table_results_std_main_paper}
  
  \vspace{-0.8em}

  \resizebox{\textwidth}{!}{

  \scriptsize

  \begin{tabular}{ c | c c c | c c c | c c | c c } 
    \toprule
    
    \multirow{2}{*} {\textbf{Method}} & \multicolumn{3}{c}{\textbf{iid}} & \multicolumn{3}{c}{\textbf{Class-iid}} & \multicolumn{2}{c}{\textbf{instance}}  & \multicolumn{2}{c}{\textbf{Class-instance}}   \\ 
    
    \cmidrule{2-11}

    & \shortstack{\textbf{iCubWorld 1.0}} & \shortstack{\textbf{CORe50}} & \shortstack{\textbf{ImageNet100}} & \shortstack{\textbf{iCubWorld 1.0}} & \shortstack{\textbf{CORe50}} & \shortstack{\textbf{ImageNet100}} & \shortstack{\textbf{iCubWorld 1.0}} & \shortstack{\textbf{CORe50}} & \shortstack{\textbf{iCubWorld 1.0}} & \shortstack{\textbf{CORe50}} \\

    \midrule 
      
      Fine-Tune & 0.1369  & 0.1145 & 0.0127 & 0.3893  & 0.3485 & 0.1233 & 0.1307  & 0.1145  & 0.3485  & 0.3430  \\
      
      EWC & - & - & - & 0.3790  & 0.3508  & 0.1225 & - & - & 0.3487  & 0.3427  \\
      
      MAS & - & - & - & 0.3912   & 0.3432  & 0.1234 & - & - & 0.3486  & 0.3429  \\
      
      VCL & - & - & - & 0.3806  & 0.3462  & 0.1205 & - & - & 0.3473  & 0.3420  \\
      
      \textit{Coreset VCL} & - & - & - & 0.3948  & 0.3424  & 0.1259 & - & - & 0.4705  & 0.4715  \\
      
      
      \textit{GDumb} & 0.8993  & 0.9345 & 0.8361 & \textbf{\textit{0.9660}} & \textit{\textbf{0.9742 }} & \textbf{\textit{0.9197}} & 0.6715  & 0.7433  & 0.7908  & 0.6548  \\
      

      DER & 0.1437  & 0.1145 & 0.0126 & 0.4057  & 0.3432 & 0.1217 & 0.3759  & 0.1168  & 0.4082  & 0.3308  \\

      DER++ & 0.1428  & 0.1145  & 0.0130 & 0.4467  & 0.3431 & 0.1230 & 0.4518  & 0.1145  & 0.4499  & 0.3429  \\
      
      TinyER & 0.9590  & 1.0007 & 0.9415 & 0.9069  & 0.9573 & 0.8995 & 0.8726  & 0.8432  & 0.8215  & 0.8461  \\
      
      ExStream & 0.9235  & 0.9844 & 0.9293 & 0.8820  & 0.8760 & 0.8757 & 0.8954  & 0.8257  & 0.8727  & 0.8837  \\
      
      
      REMIND & 0.9260  & 0.9933 & 0.9088 & 0.8553  & 0.9448 & 0.8803 & 0.8157  & 0.8544  & 0.7615  & 0.7826  \\
      
      
      \textbf{Ours} & \textbf{0.9716 } & \textbf{1.0069 } & \textbf{0.9640} & \textbf{0.9480 } & \textbf{0.9686 } & \textbf{0.9171} & \textbf{0.9580 } & \textbf{0.9824 } & \textbf{0.9585 } & \textbf{0.9384 } \\
     
    \midrule
     
    Offline & 1.0000 & 1.0000 & 1.0000 & 1.0000 & 1.0000 & 1.0000 & 1.0000 & 1.0000 & 1.0000 & 1.0000  \\
     
     
    $\widehat{\text{Offline}}$ & 0.7626 & 0.8733 & 0.8520 & 0.8849 & 0.9070 & 0.8953 & 0.7646 & 0.8733 & 0.8840 & 0.9079 \\ 
     
   \bottomrule
   
  \end{tabular}

  }

  \vspace{-2.2em}
  
\end{table*}

\textbf{Datasets.} To evaluate the efficacy of the proposed model we perform extensive experiments on three benchmark datasets: ImageNet100, iCubWorld 1.0~\cite{fanello2013icub}, and CORe50~\cite{pmlr-v78-lomonaco17a}. ImageNet100 is a subset of ImageNet-1000 (ILSVRC-2012)~\cite{russakovsky2015imagenet} containing randomly chosen 100 classes, with each class containing 700-1300 training samples and 50 validation samples. Since the test data for ImageNet-1000 (ILSVRC-2012)~\cite{russakovsky2015imagenet} is not provided with labels, we use the validation data for evaluating the model's performance, similar to \cite{hayes2019remind}. iCubWorld 1.0 is an object recognition dataset which contains the sequences of video frames, with each containing a single object. There are 10 classes, each containing 3 different object instances with each class containing 600-602 samples for training and 200-201 samples for testing. CORe50 is similar to iCubWorld 1.0, containing images from temporally coherent sessions. There are 10 classes, each containing 5 different object instances with each class containing 11983-12000 samples for training and 4495-4500 samples for testing. \emph{Technically, iCubWorld, and CORe50 are the ideal dataset for streaming learning, as it requires learning from temporally coherent image sequences, which are naturally non-i.i.d images}.

\textbf{Evaluation Over Different Data Orderings.} The proposed approach is robust to the various \emph{SL} settings; we evaluate the model's SL ability with the following four~\cite{hayes2019memory, hayes2019remind} challenging data ordering schemes: $(i)$ \emph{`streaming iid'}: where the data-stream is organized by the randomly shuffled samples from the dataset, $(ii)$ \emph{`streaming class iid`}: where the data-stream is organized by the samples from one or more classes, these samples are shuffled randomly, $(iii)$ \emph{`streaming instance'}: where the data-stream is organized by temporally ordered samples from different object instances, and $(iv)$ \emph{`streaming class instance'}: where the data-stream is organized by the samples from different classes, the samples within a class are temporally ordered based on different object instances. \emph{Only iCubWorld 1.0 and CORe50 dataset contains the temporal ordering, therefore `streaming instance', and `streaming class instance' settings are evaluated only on these two datasets.}


\subsection{Metrics}\label{metrics}

For evaluating the performance of the streaming learner, we use $\boldsymbol{\Omega_{\text{all}}}$ metric, similar to \cite{kemker2018measuring, hayes2019memory, hayes2019remind}, where $\boldsymbol{\Omega}_{\text{all}}$ represents normalized incremental learning performance with respect to an offline learner: $\boldsymbol{\Omega}_{\text{all}} = \frac{1}{T} \sum_{t = 1}^{T} \frac{\boldsymbol{\alpha}_{t}}{\boldsymbol{\alpha}_{\text{offline}, t}}$, where $T$ is the total number of testing events, $\boldsymbol{\alpha}_{t}$ is the performance of the incremental learner at time $t$, and $\boldsymbol{\alpha}_{\text{offline}, t}$ is the performance of a traditional offline model at time $t$.



\subsection{Baselines And Compared Methods}\label{baselines_and_compared_methods}

The proposed approach follows the \emph{`streaming learning setup'}; to the best of our knowledge, recent works ExStream~\cite{hayes2019memory} and REMIND~\cite{hayes2019remind} are the only methods that follow the same setting. We compare our approach against these strong baselines. We also compare our model with $(i)$ a network trained with one sample at a time (Fine-tuning/lower-bound) and $(ii)$ a network trained offline, assuming all the data is available (Offline/upper-bound). Finally, we choose recent popular `batch' (IBL) and `online' learning methods, such as EWC~\cite{kirkpatrick2017overcoming}, MAS~\cite{aljundi2018memory}, VCL~\cite{nguyen2017variational}, Coreset VCL~\cite{nguyen2017variational}, TinyER~\cite{chaudhry2019tiny}, GDumb~\cite{prabhu2020gdumb}, and DER/DER++~\cite{buzzega2020dark} as baselines and rigorously evaluate our model against these approaches. For a fair comparison, we train all the methods in a SL setup, i.e., one sample at a time, and no finetuning is used. Only \textit{Coreset VCL} and \textit{GDumb} finetunes before each inference, therefore, these two methods utilize an extra advantage over the other baselines. However, {\pname} still outperforms these approaches by a significant margin.




\begin{figure}[ht]
  \centering
  \vspace{-0.5em}
  \includegraphics[width=8.5cm, height=2.7cm]{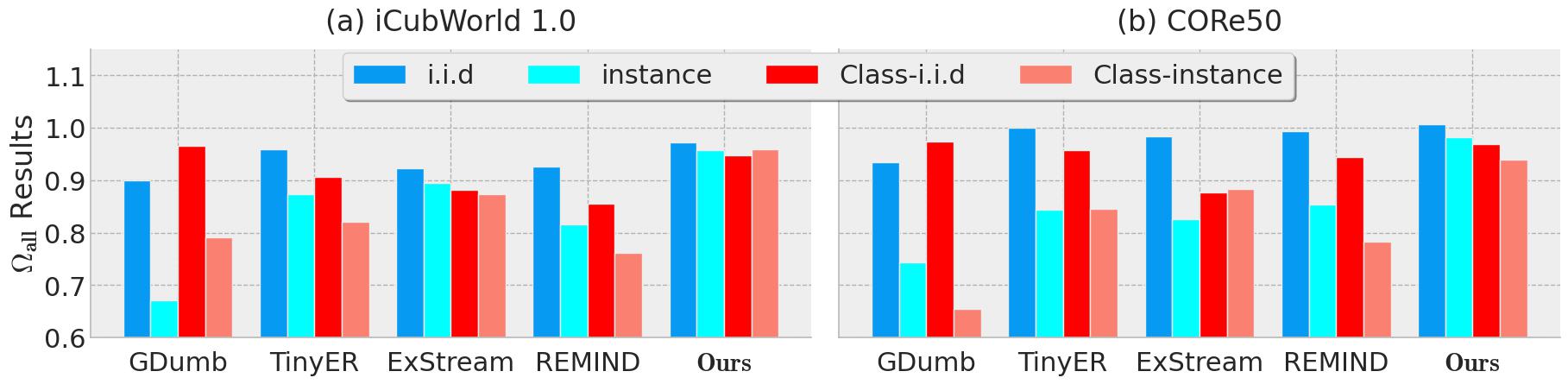}
  \vspace{-0.5em}
  \caption{Plots of $\boldsymbol{\Omega}_{\text{all}}$ as a function of streaming learning model and data-ordering on $(a)$ iCubWorld 1.0, and $(b)$ CORe50.}
  

  \vspace{-1.2em}
  \label{fig:temporal_coherance}
\end{figure}

\begin{figure*}[ht]
    \centering

    
    \includegraphics[width=14cm, height=2.5cm]{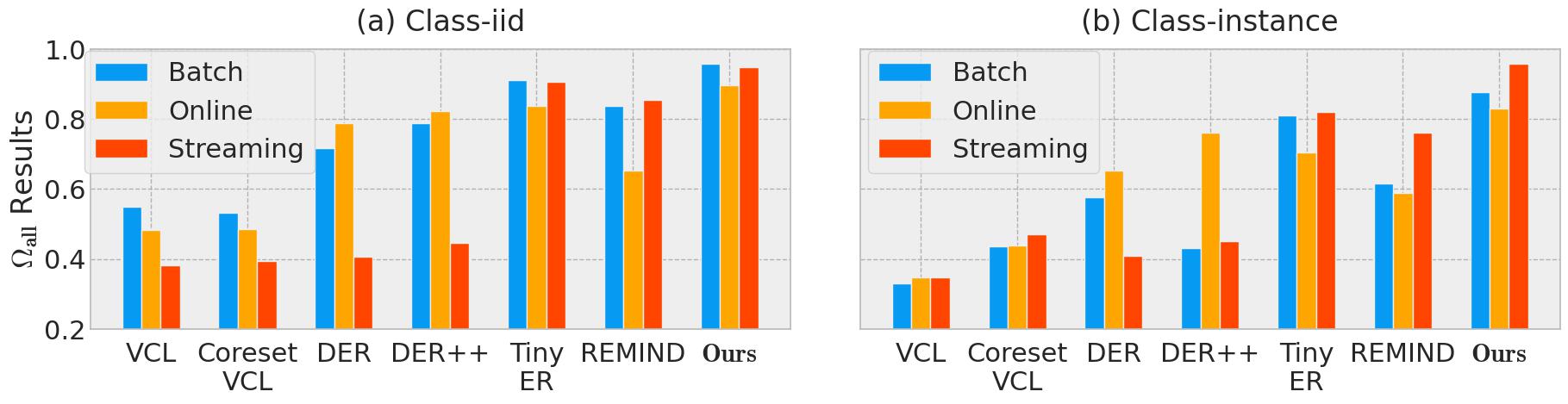}

    \vspace{-0.5em}

    \caption{$\boldsymbol{\Omega}_{\text{all}}$ results for different versions of baselines on iCubWorld 1.0 on $(a)$ Class-iid, and $(b)$ Class-instance ordering.}
    \label{fig:batch_vs_online_vs_streaming} 
    
    \vspace{-1.8em}
\end{figure*}

\subsection{Results}\label{results}


The detailed results of {\pname} over various experimental settings along with the strong baseline methods are shown in Table~\ref{Table_results_std_main_paper}. We can clearly observe that {\pname} consistently outperforms all the baseline by a significant margin. The proposed model is also robust to the different streaming learning scenarios compared to the baselines. We repeat our experiment ten times and report the average accuracy. We observe that `batch-learning' methods severely suffer from catastrophic forgetting. Moreover, replay-based `online-learning' methods such as DER/DER++ also suffer from information loss badly.

While \textit{GDumb} achieves higher final accuracy on class-i.i.d ordering, it finetunes before each inference. Therefore, it enjoys an extra advantage over other SL baselines and it is not considered as the best-performing model, even when it achieves higher accuracy.


We believe it is important to highlight that iCubWorld 1.0 and CORe50 are two challenging datasets, which evaluate the models in more realistic scenarios or data-orderings. Particularly, class-instance and instance ordering require the learner to learn from temporally ordered video frames one at a time. From Table~\ref{Table_results_std_main_paper}, we observe that {\pname} obtain up to ${8.58}\%$ \& ${6.26}\%$ improvement on iCubWorld 1.0, and ${5.47}\%$ \& ${12.8}\%$ improvement on CORe50, over the state-of-the-art streaming learning approaches. Fig.~\ref{fig:temporal_coherance} shows the impact of temporal orderings on the streaming learning model's performance. It is evident that class-instance and instance ordering are more difficult, and the baselines continue to suffer from severe forgetting. 


Finally, for completeness, we train {\pname} in `batch' as well as `online' learning setting to determine its effectiveness and compatibility in these settings. In Fig.~\ref{fig:batch_vs_online_vs_streaming}, we compare {\pname} in various settings. We can observe that {\pname} outperforms the baselines by a significant margin on both class-i.i.d and class-instance ordering on iCubWorld, \emph{implying even though {\pname} designed to work in SL, it can be thought of as a robust method for various LLL scenarios with the widest possible applicability}. 



\subsection{Implementation Details}\label{implementation_details}

In all the experiments, models are trained with one sample at a time. Only \textit{Coreset VCL} and \textit{GDumb} finetunes before each inference. For a fair comparison, the same network structure is used throughout all the models. For all methods, we use fully connected single-head networks with two hidden layers as the plastic network $F(\cdot)$, where each layer contains $256$ nodes with ReLU activations; for `VCL', `Coreset VCL', and {`{\pname}'}, $F(\cdot)$ is a BNN, whereas for all other methods $F(\cdot)$ is a deterministic network. For a fair comparison, we store the same number of past examples for all replay-based approaches, as mentioned in Table~\ref{Table_buffer_capacity}. We use Mobilenet-V2~\cite{sandler2018mobilenetv2} pre-trained on ImageNet-1000 (ILSVRC-2012)~\cite{russakovsky2015imagenet} as the visual feature extractor. For memory-replay, we use \emph{`uncertainty-aware positive-negative'} sampling strategy throughout all data-orderings, except for `streaming-i.i.d' ordering, we use \emph{`uniform'} sampling. We use \emph{`loss-aware weighted random replacement with a reservoir'} sampling strategy as memory replacement policy for all the experiments. We set $N_{1}^{\prime} = N_{2}^{\prime} = 16$ throughout all experiments. We set $\lambda_{1}= 1$ and $\lambda_{2} = 0.3$ across all experiments; however, for \textit{online/batch} learning experiments, we use $\lambda_{2} = 0.2$ and use \emph{uniform sampling} for memory replay. We repeat our experiment $10$ times and report the average-accuracy.

\section{Ablation Study}\label{ablation_study}

\vspace{-0.5em}

\begin{table}[t]
  \scriptsize
  \centering

  
  \caption{$\boldsymbol{\Omega}_{\text{all}}$ Results as a function of different sampling and buffer management policies.}
  
  \label{Table_different_sampling_ablation}
  
  
  \begin{tabular}{ c | c | c | c | c | c } 
    \toprule

     \multirow{3}{*} {\shortstack{\textbf{Memory} \\ \textbf{Replacement}} } & \multirow{3}{*} {\shortstack{\textbf{Sample} \\ \textbf{Selection}}} & \multicolumn{2}{c}{\textbf{iCubWrold}} & \multicolumn{2}{c}{\textbf{ImageNet100}} \\
    
    \cmidrule{3-6}
   &  & \shortstack{\textbf{instance}} & \shortstack{\textbf{Class} \\ \textbf{instance}} & \shortstack{\textbf{iid}} & \shortstack{\textbf{Class-iid}}  \\
    \midrule

     \multirow{3}{*} { LAWCBR }  & Uni & 0.8975  & 0.8506 & 0.9582   & 0.9014  \\
     
    & UAPN & 0.9346 & 0.8500 &  0.9327  & 0.9135   \\
     
    & LAPN & 0.9172  & 0.8536 &  0.9253  & 0.9122   \\

    \cmidrule{1-6}

    \multirow{3}{*} { LAWRRR }  & Uni & 0.9269  & 0.9346 &  \textbf{0.9640}  & 0.8643  \\
     
    & UAPN & \textbf{0.9580} & \textbf{0.9585 } &  0.9578 & \textbf{0.9171}  \\
     
    & LAPN & 0.9558 & 0.9497 & 0.9575 & 0.9112  \\


     
     
     
   \bottomrule
  \end{tabular}
  
  \vspace{-0.4em}
  
\end{table}

We perform extensive ablation to validate the significance of the proposed components.



\textbf{Significance Of Different Sampling Strategies.} In Table~\ref{Table_different_sampling_ablation}, we compare the performance of {\pname} while using various sampling strategies and memory replacement policies. We observe that for the buffer replacement, LAWRRR performs better compared to LAWCBR. Furthermore, for the sample replay, UAPN, along with LAWRRR memory buffer policy, outperforms other sampling strategies, except \emph{uniform sampling} (Uni) performs better on i.i.d ordering. 



\begin{figure}[!htp]
  \centering
  \vspace{-0.5em}
  \includegraphics[width=8.2cm, height=2.8cm]{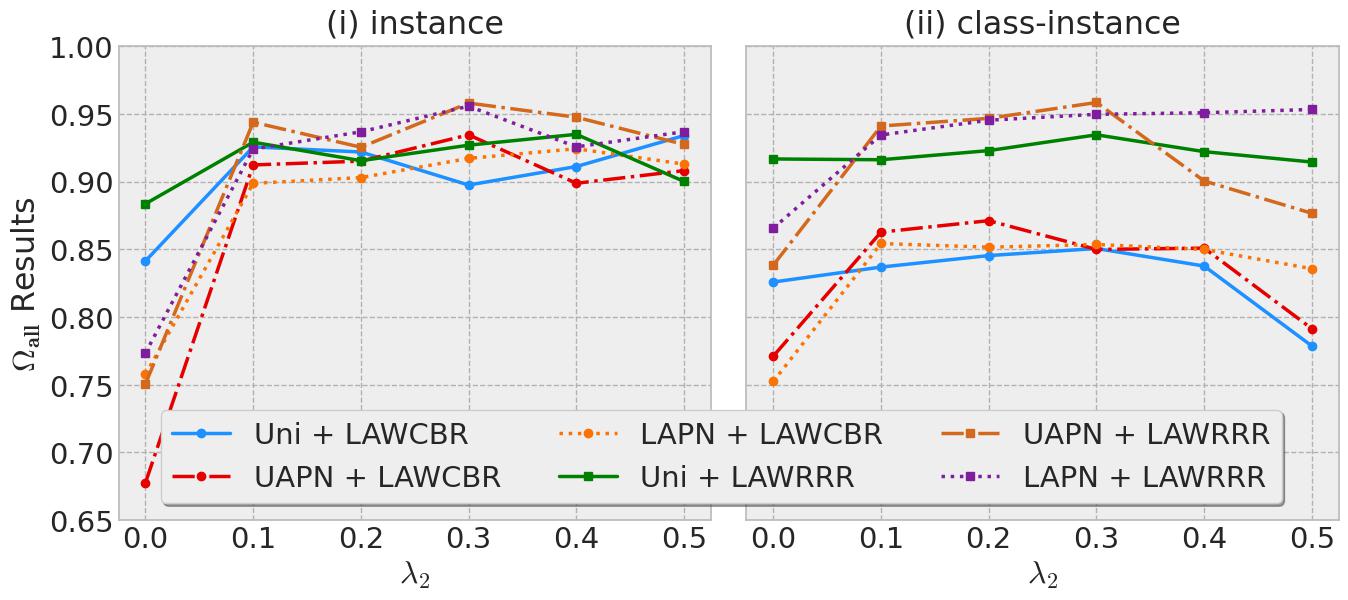}
  \vspace{-0.5em}
  \caption{Plots of $\boldsymbol{\Omega}_{\text{all}}$ as a function of hyper-parameter $\boldsymbol{\lambda}_{2}$ and different sampling strategies and replacement policies for $(i)$ instance, $(ii)$ class-instance ordering on iCubWorld 1.0.} 

  \vspace{-1.0em}
  \label{fig:lambda_2_ablation_iCubWorld}
\end{figure}

\textbf{Choice Of Hyperparameter ( $\boldsymbol{\lambda}_{2}$).} Fig.~\ref{fig:lambda_2_ablation_iCubWorld} shows the effect of changing the knowledge-distillation loss weight $\boldsymbol{\lambda}_{2}$ on the final $\boldsymbol{\Omega}_{\text{all}}$ accuracy for iCubWorld 1.0 on instance and class-instance ordering, while using different sampling strategies and buffer replacement policies. We observe the best model performance for $\boldsymbol{\lambda}_{2} = 0.3$, and use this value for all our experiments. 



\textbf{Significance Of Knowledge-Distillation Loss.}  Fig.~\ref{fig:lambda_2_ablation_iCubWorld} with $\boldsymbol{\lambda}_{2} = 0.0$ represents the model without kd-loss. We can observe that the model performance significantly degrades without knowledge distillation. Therefore, kd-loss is a key component to the model's performance. 


\begin{figure}[!htp]
    \centering
    \includegraphics[width=6.5cm, height=2.8cm]{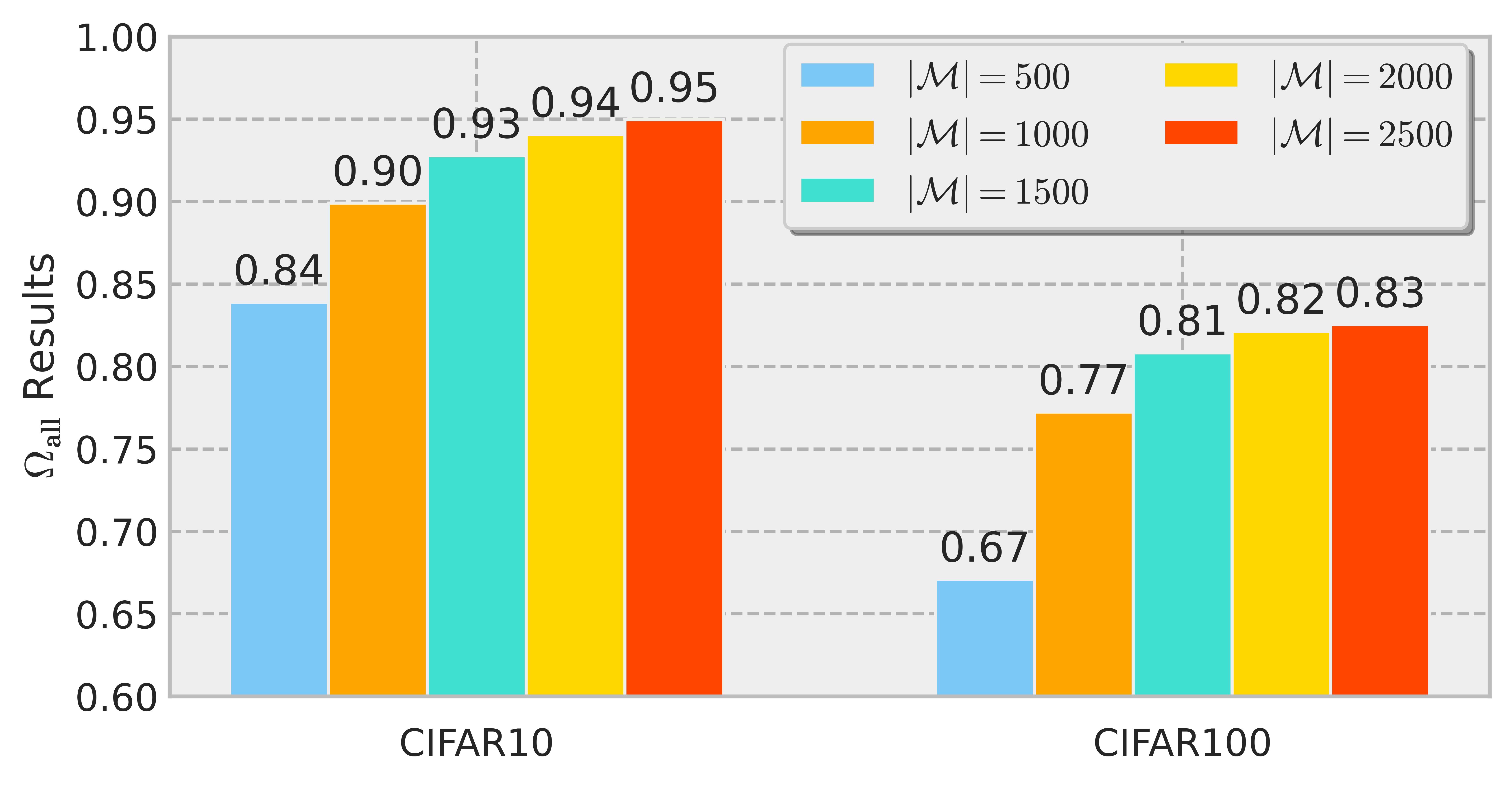}
    \vspace{-0.5em}
    \caption{Plots of $\boldsymbol{\Omega}_{\text{all}}$ as a function of buffer capacity $|\boldsymbol{M}|$ for class-i.i.d data-ordering on CIFAR10 and CIFAR100.} 
    \label{fig:memory_ablation_study}
    \vspace{-1.2em}
\end{figure}

\textbf{Choice Of Buffer Capacity.} We perform an ablation for the different buffer capacities, i.e., $|\boldsymbol{M}|$. The results are shown in Figure~\ref{fig:memory_ablation_study}. It is evident that, with the longer sequence of incoming data, the model's ({\pname}) performance improves with the increase in the buffer capacity, as it helps minimize the confusion in the output prediction.

\vspace{-0.2em}

\section{Conclusion}\label{conclusion}

\vspace{-0.5em}

\emph{Streaming continual learning} (SCL) is the most challenging and realistic framework for CL; most of the recent promising models for CL are unable to handle this above setting. Our work proposes a dual regularization and loss-aware buffer replacement to handle the SCL scenario. The proposed model is highly efficient since it learns a joint likelihood from the current and replay samples without leveraging any external finetuning. Also, to improve the training efficiency further, the proposed model selects a few most informative samples from the buffer instead of using the entire buffer for the replay. We have conducted a rigorous experiment over several challenging datasets and showed that {\pname} outperforms the recent state-of-the-art approaches in this setting by a significant margin. To disentangle the importance of the various components, we perform extensive ablation studies and observe that the proposed components are essential to handle the SCL setting.

\medskip
{\small
\bibliographystyle{IEEEtran}
\bibliography{main_paper.bib}
}

\end{document}